\documentclass[11pt]{article}

\usepackage[preprint]{acl}

\usepackage{times}
\usepackage{latexsym}

\usepackage[T1]{fontenc}

\usepackage[utf8]{inputenc}

\usepackage{microtype}

\usepackage{inconsolata}

\usepackage{graphicx}
\usepackage{multirow}
\usepackage{booktabs}
\usepackage[table]{xcolor}
\usepackage{pifont}
\usepackage{makecell}
\usepackage{listings}
\definecolor{dkgreen}{rgb}{0,0.6,0}
\definecolor{gray}{rgb}{0.5,0.5,0.5}
\definecolor{mauve}{rgb}{0.58,0,0.82}

\title{MemTools: A Unified Research Framework for Interoperable Agent Memory}

\author{
 \textbf{Chengfeng Zhao\textsuperscript{1,2}},
 \textbf{Jinhui Chen\textsuperscript{1,2}},
 \textbf{Sirui Liang\textsuperscript{1,2,4}},
 \textbf{Shizhu He\textsuperscript{1,2}},
 \textbf{Yequan Wang\textsuperscript{3}},
 \textbf{Jun Zhao\textsuperscript{1,2}},
 \textbf{Kang Liu\textsuperscript{1,2}\thanks{Corresponding Author}}
\\
\\
 \textsuperscript{1}Institute of Automation, CAS
 \textsuperscript{2}University of Chinese Academy of Sciences
\\
 \textsuperscript{3}Beijing Academy of Artificial Intelligence
 \textsuperscript{4}Zhongguancun Institute of Artificial Intelligence
\\
\texttt{\{zhaochengfeng2024, chenjinhui2025,liangsirui2024\}@ia.ac.cn} \\
\texttt{\{shizhu.he, kliu, jzhao\}@nlpr.ia.ac.cn},
\texttt{tshwangyequan@gmail.com} \\
}

\begin{document}
\maketitle
\begin{abstract}
While memory systems are essential for agent architectures, pervasive architectural fragmentation restricts systematic research.
Existing implementations typically couple different stages of the memory lifecycle, entangle evaluation logic with specific datasets, and provide limited support for the management of heterogeneous memory types.
We introduce MemTools, an interoperability research framework that decouples memory system components from their underlying deployment environments. MemTools standardizes the memory lifecycle through declarative data contracts, enabling the interchangeable assembly of components across different systems. It orthogonally separates benchmark datasets from execution protocols to facilitate controlled assessments. 
Furthermore, MemTools provides a unified computational interface for coordinating symbolic, neural, and multimodal memory representations within a shared runtime.
Empirical evaluations on cross-system component integration, evaluation protocol reconfiguration, and heterogeneous memory coordination demonstrate that MemTools enables systematic isolation and analysis of memory design variables. These findings suggest that MemTools provides a practical and extensible infrastructure for advancing principled research on agent memory.
\end{abstract}

\section{Introduction}

Large language model agents increasingly rely on long-term memory to manage complex tasks and sustain operations over long horizons~\cite{hu2025memory, li2025externalization}. While numerous memory systems have been developed recently~\cite{memgpt, zep, memos, mem0, mirix, omni-simplemem, m3agent}, systematic research remains difficult due to significant variation in system architectures~\cite{luo2026storageexperiencesurveyevolution,huang2025rethinking}. Existing memory implementations frequently couple core components with specific deployment environments and setups. This coupling prevents researchers from isolating individual design variables and makes it difficult to identify the underlying sources of performance differences across architectures.

This architectural entanglement occurs at multiple levels. As summarized in Table~\ref{tab:comparison}, current systems share the following limitations: 1) \textbf{Coupled Lifecycle Stages}:
At the component level, lifecycle stages such as memory formation, storage, retrieval, and evolution are typically encapsulated within closed codebases~\cite{MemEngine,tang2026unifiedrepresentation}.
The absence of declarative data contracts restricts the reuse of individual components across different systems.
2) \textbf{Entangled Evaluation Protocols}: At the evaluation level, memory pipelines are often bound to specific testing frameworks. Since performance variations frequently reflect differences in execution logic rather than underlying memory capabilities, entangling datasets with evaluation protocols introduces assessment bias~\cite{huang2025rethinking,wang2026evomembenchbenchmarkingagentmemory}.
3) \textbf{Fragmented Heterogeneous Memory}: At the representation level, the lack of standardized interfaces makes it challenging to coordinate heterogeneous memory formats, such as symbolic databases, neural parameters, and multimodal records, within a single experimental runtime.

\begin{table*}
\centering

\newcommand{\yes}{\textcolor{green}{\ding{108}}}
\newcommand{\partly}{\textcolor{yellow}{\ding{108}}} 
\newcommand{\no}{\textcolor{red}{\ding{108}}}

\begin{tabular}{l c c c c c c}
\toprule[2pt]
\multirow{2}{*}{System}
    & \multicolumn{2}{c}{Lifecycle}
    & \multicolumn{1}{c}{Scenario}
    & \multicolumn{3}{c}{Representation} \\
    \cmidrule(lr){2-3} \cmidrule(lr){4-4} \cmidrule(lr){5-7}
    & \thead{Stage\\Separation}
    & \thead{Cross-system\\Interoperability}
    & \thead{Scenario\\Coverage}
    & \thead{Symbolic}
    & \thead{Neural}
    & \thead{Multi-\\modal} \\
\midrule
MemGPT~\cite{memgpt}          & \no     & \no     & \no     & \partly & \no     & \no     \\
Zep~\cite{zep}             & \partly & \no     & \no     & \yes    & \no     & \no     \\
MemOS~\cite{memos}           & \yes    & \partly & \partly & \yes    & \yes    & \partly \\
Mem0~\cite{mem0}            & \partly & \partly & \no     & \yes    & \no     & \no     \\
MIRIX~\cite{mirix}           & \no     & \no     & \no     & \partly & \no     & \partly \\
Omni-SimpleMem~\cite{omni-simplemem}  & \partly & \no     & \no     & \yes    & \no     & \yes    \\
M3-Agent~\cite{m3agent}        & \partly & \no     & \no     & \partly & \no     & \yes    \\
MemFactory~\cite{MemFactory}        & \yes & \partly     & \partly     & \yes & \no     & \no    \\
MemEngine~\cite{MemEngine}        & \yes & \partly     & \no     & \partly & \no     & \no    \\
ReMe~\cite{ReMe}        & \yes & \partly     & \no     & \yes & \no     & \no    \\
\textbf{MemTools (Ours)} & \yes    & \yes    & \yes    & \yes    & \yes    & \yes    \\
\bottomrule[2pt]
\end{tabular}
\caption{Comparison of different agent memory architectures across lifecycle standardization, scenario unification, and representational diversity. \yes~= full support; \partly~= partial support; \no~= no or limited support. \textbf{Stage Separation} evaluates the independence of module boundaries within the memory lifecycle. \textbf{Cross-system Interoperability} denotes the feasibility of integrating components from distinct architectures. \textbf{Scenario Coverage} reflects support for different memory application scenarios and protocols. \textbf{Symbolic/Neural/Multimodal} denotes the specific memory representations supported by each system.}
\label{tab:comparison}
\vspace{-10pt}
\end{table*}

To address these limitations, we present MemTools, an interoperability research framework designed to decouple agent memory pipelines from their deployment environments. 
Specifically, MemTools has the following advantages: 1) \textbf{Standardized Lifecycle Interface}:
MemTools standardizes the memory lifecycle by defining declarative data contracts, allowing researchers to swap and combine components from different architectures, including memory backend, formation, retrieval, evolution, and utilization. A matching engine automatically verifies compatibility and enumerates valid pipeline combinations by checking provided and required data fields. This enables the interchangeable assembly of components across different systems without manual code integration.
2) \textbf{Unified Evaluation Paradigms}:
The framework orthogonally decouples benchmark datasets from evaluation protocols. Implementations span both user-centric and agent-centric scenario datasets and support a suite of customizable protocols. By abstracting execution logic into pluggable protocol interfaces, a single dataset can be processed under multiple evaluation workflows, effectively isolating the intrinsic memory capabilities from setup-induced biases.
3) \textbf{Heterogeneous Memory Management}:
MemTools supports heterogeneous memory representations by coordinating symbolic, neural, and multimodal memory under a unified computational interface. A multisystem coordination layer manages these distinct pipelines concurrently using synchronized parallel search and chained utilization.

We demonstrate the practical utility of MemTools through several empirical studies. 
By integrating diverse existing systems into our standard interfaces, we construct hybrid memory pipelines, manipulate evaluation logic independently of task datasets, and coordinate disparate memory modalities concurrently.
These controlled experiments reveal that cross-system component assemblies can match or exceed native designs, isolating evaluation logic reveals the critical impact of memory operation timing on performance, and that heterogeneous memory coordination provides complementary performance gains.
Overall, these findings establish MemTools as an effective and versatile framework for researchers systematically investigating agent memory. The code is open-source at \url{https://github.com/JJJAYYYZhao/MemTools-public}. We also provide a demonstration video on \url{https://youtu.be/nH03SeVBOPc}.
\begin{figure*}[t!]
    \centering
    \includegraphics[width=0.95\textwidth]{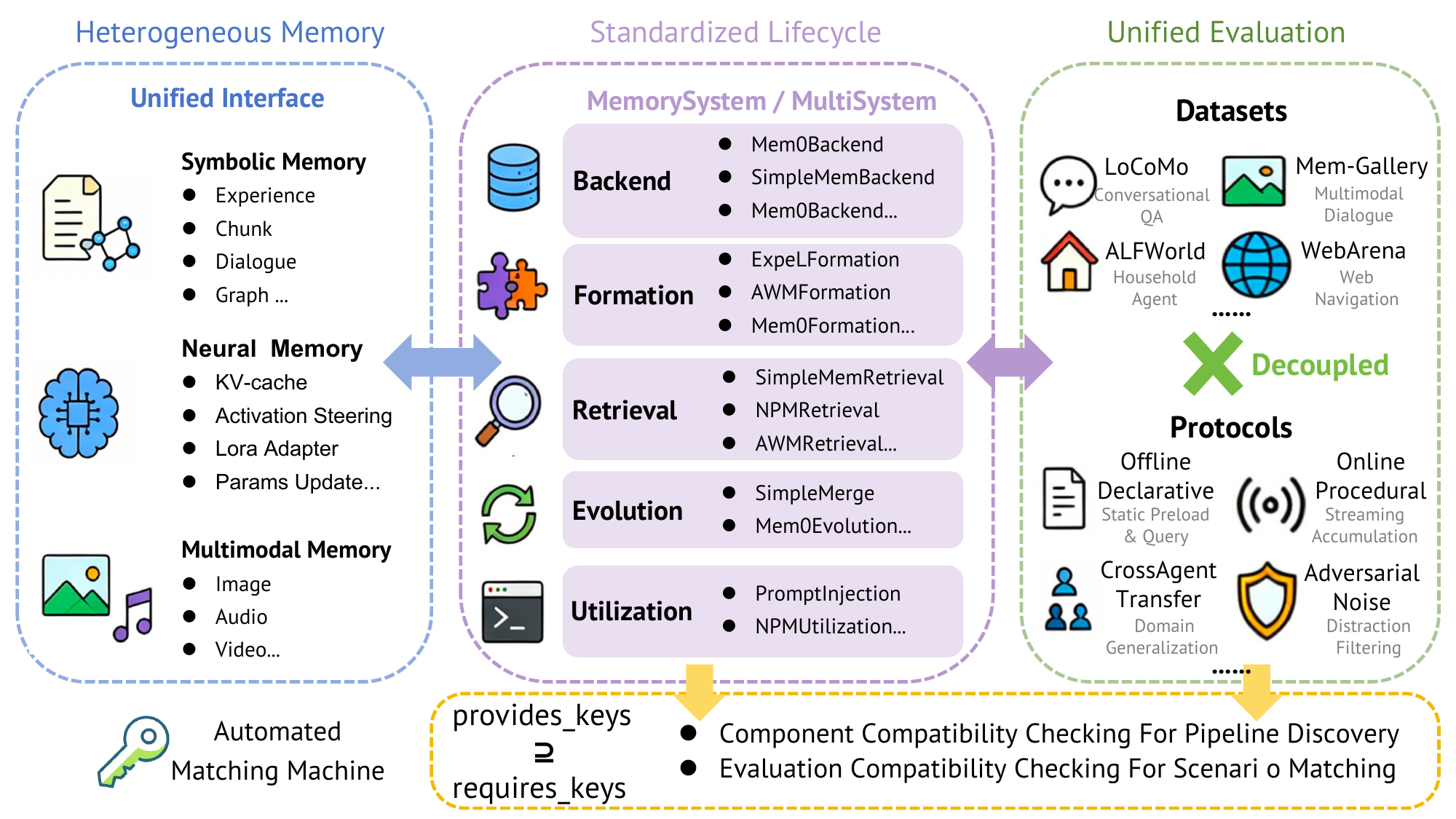}
    \caption{The MemTools architecture. Heterogeneous memory coordination unifies symbolic, neural, and multimodal memory using a unified interface (left). The core MemorySystem pipeline standardizes the memory lifecycle with different components (center).
    Benchmarks and evaluation protocols are decoupled through an orthogonal design (right).  The key matching engine verifies component compatibility automatically (bottom).}
    \vspace{-10pt}
    \label{fig:architecture}
\end{figure*}

\vspace{-15pt}
\section{Related Work}
\vspace{-5pt}
\paragraph{Memory System Implementations}
Recent studies characterize agent memory as a lifecycle of distinct functional stages~\cite{surveyonmemorymechanism,hu2025memory,huang2025rethinking} and propose various methods to sustain long-term memory capabilities~\cite{zheng2023synapse,memorybank, gutierrez2024hipporag, han2025legomemmodularproceduralmemory, lightmem, ouyang2025reasoningbankscalingagentselfevolving}.
However, the implementations of these methods remain tightly coupled to their host architectures.
The lack of standardized data contracts between stages prevents the extraction and reuse of independent components. This architectural rigidity hinders systematic comparative research across different systems.

\paragraph{Memory Evaluation Protocols and Datasets}
The assessment of memory systems is generally divided by application domains. User-centric evaluations measure factual recall and persona consistency via offline question-answering loops~\cite{locomo,wu2025longmemevalbenchmarkingchatassistants,zhang2025personalization, liu2025echolargelanguagemodel,xu2026amem}. Agent-centric evaluations track continuous trajectories in tasks requiring numerous reasoning and action steps~\cite{expel,awm,gutierrez2025from,ouyang2025reasoningbankscalingagentselfevolving}, such as web navigation~\cite{WebShop,webarena}, coding~\cite{jimenez2024swebench}, or multi-turn decision-making~\cite{chevalier-boisvert2018babyai,alfworld,wang2022scienceworld}. However, the execution logic governing memory operations is usually bound to the specific benchmark designs. Isolating the evaluation protocols from benchmark datasets is necessary to observe how different memory timing strategies affect algorithmic performance.

\paragraph{Memory Representations}
Agent memory utilizes several distinct representations. Symbolic memories organize interaction histories into vector databases or relational graphs~\cite{mem0,zep,mirix}. Neural approaches encode information directly into model weights or continuous hidden states~\cite{MEMORYLLM,wang2025m+,wang2025selfupdatable,MemGen,yu2026latentskillincontexttextualskills}. Multimodal memories store sensory inputs like images or audio~\cite{m3agent,omni-simplemem}. Existing frameworks usually specialize in a single memory representation. Coordinating symbolic, neural, and multimodal memory within a runtime environment requires a shared computational interface.

\section{System Design}

Figure~\ref{fig:architecture} illustrates the MemTools architecture. The framework refactors the memory lifecycle into discrete stages governed by declarative data contracts, orthogonally decouples evaluation protocols from benchmark datasets, and coordinates symbolic, neural, and multimodal memory representations as parallel subsystems under a unified interface.

\subsection{Standardized Lifecycle Interface}

MemTools structures the memory lifecycle into four stages and a shared storage backend. The framework establishes public typed interfaces for formation, retrieval, evolution, and utilization. Data flows between these discrete components via a standardized \verb|MemoryEntry| data structure. This uniform carrier ensures that outputs from one stage are correctly interpreted by the next, regardless of their underlying implementations.

To maintain stability during component integration, components communicate through declarative data contracts. Each component explicitly declares the data fields it requires to function (\verb|requires_keys|) and the fields it provides upon execution (\verb|provides_keys|). The framework validates pipeline compatibility during initialization by confirming that the fields provided by upstream components fulfill the dependencies of downstream components. For example, a retrieval strategy expecting a dense embedding index will only pair with a storage backend capable of generating that specific format.  Operating on these data contracts, a matching engine automatically discovers and enumerates all valid pipeline combinations without manual configuration, seamlessly supporting baseline strategies, adapters for existing systems, and custom components.

\begin{figure}[t]
\centering
\includegraphics[width=\linewidth]{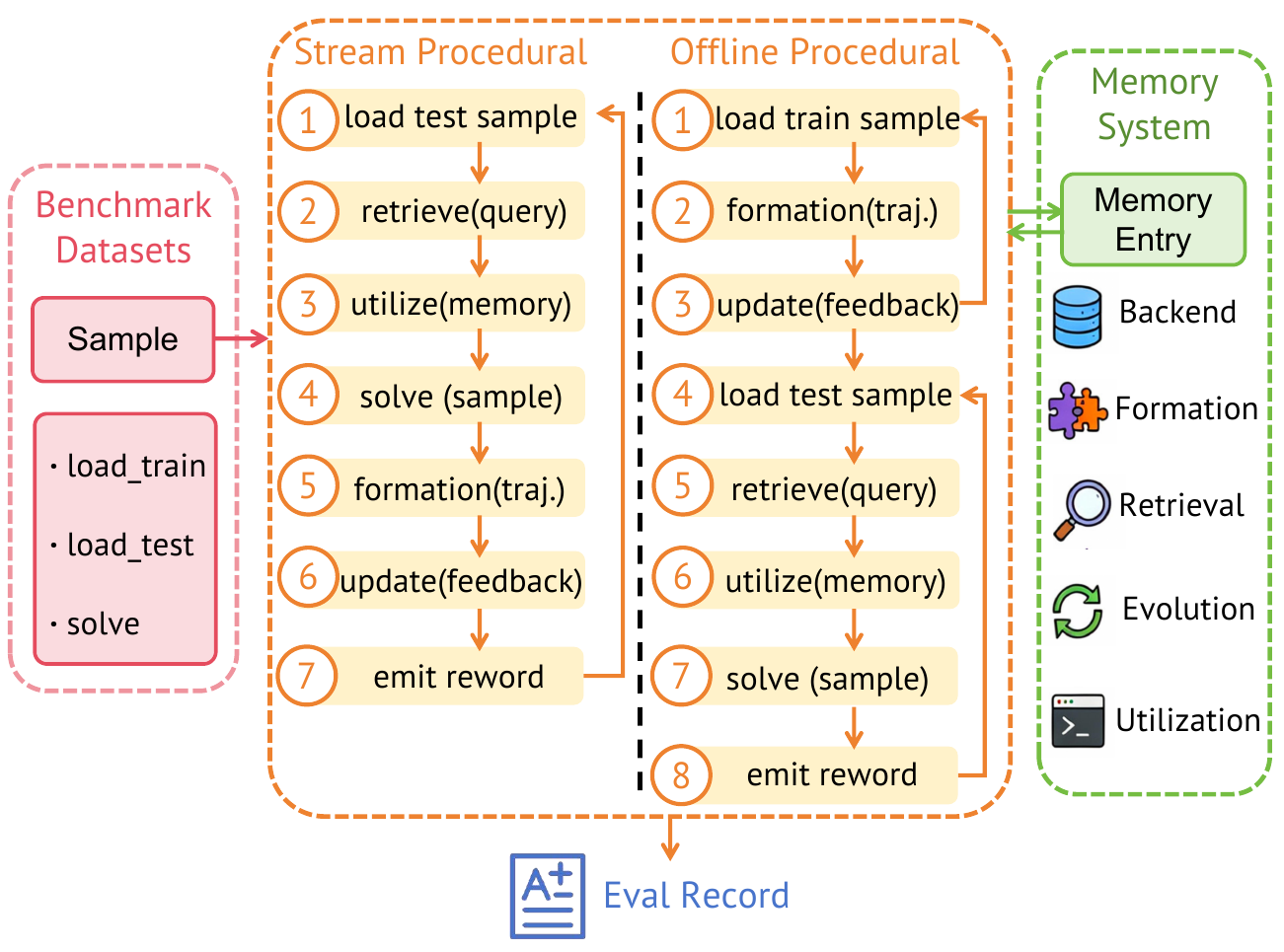} 
\caption{Data flows of different evaluation protocols.}
\vspace{-10pt}
\label{fig:workflow}
\end{figure}

\subsection{Unified Evaluation Paradigms}
To facilitate fair comparative analysis, the framework abstracts execution logic into pluggable evaluation protocols that dictate the sequence of memory extraction, retrieval, adaptation, and task solving. As illustrated in Figure~\ref{fig:workflow}, 
within this architecture, benchmarks function strictly as data providers, encapsulating raw information into standardized \verb|Sample| objects. This orthogonal design enables a single dataset to be processed under multiple evaluation workflows. The data contract mechanism similarly governs this stage, verifying that a benchmark supplies the exact fields required by a chosen protocol before execution begins.

The framework supports diverse evaluation scenarios through configurable execution flows. For static environments, declarative protocols handle conversational tasks by extracting knowledge from reference corpora before executing isolated search steps. In dynamic agent environments, procedural protocols evaluate memory through continuous trajectory accumulation, ranging from offline trajectory collection to online paradigms that interleave memory formation with task execution. By treating these execution timings as configurable parameters rather than hardcoded logic, the architecture unifies previously fragmented assessment methodologies.

\subsection{Heterogeneous Memory Management}
Diverse memory representations necessitate fundamentally different operational logic. MemTools encapsulates each representation into an independent processing pipeline. Every pipeline strictly adheres to the standardized lifecycle interface, allowing the framework to manage underlying technologies through identical operational calls. A \verb|MultiSystem| coordination layer acts as a unified interface, abstracting the complexity of heterogeneous backends away from the base language model. 

This coordination layer orchestrates the integration of parallel pipelines through synchronized search operations and sequential utilization. When an agent initiates a query, the layer dispatches the request concurrently across all active memory subsystems, aggregating relevant textual records, visual contexts, and neural activation patterns. During the utilization stage, these retrieved components are applied to the language model through a chained adaptation process. Consequently, researchers can seamlessly expand this ecosystem to support new memory representations by constructing pipelines that follow the established lifecycle contracts.

\subsection{Framework Usage}

Figure~\ref{fig:code_snippet} illustrates the typical workflow of MemTools through its Python API. 
Developers begin by loading a benchmark dataset and configuring the base language model using the shared \texttt{LLM} interface. 
They then assemble a \texttt{MemorySystem} pipeline by plugging specific components into the lifecycle stages.
Since these components conform to the standardized data contracts, researchers can combine built-in strategies with adapters from existing architectures. 
Finally, the evaluation protocol is selected and executed via the \texttt{run()} method, which manages memory operations throughout the task trajectory and outputs structured evaluation records. 
This code structure demonstrates the design logic of the framework, where the benchmark data, the evaluation logic, and the memory implementation operate as strictly independent variables during experiments.

\begin{figure}[ht]
    \centering
    \begin{lstlisting}
# Import core API, benchmark datasets, protocols, and modular components (omitted for brevity) ...

# Step 1: Load benchmark and configure LLM
benchmark = ALFWorldBenchmark(data_path="...")
llm = LLM(model_name="xxx")

# Step 2: Assemble memory pipeline from pluggable components
ms = MemorySystem(
    backend=InMemoryStore(), 
    retrieval=BM25Search(), # built-in component
    formation=ExpelFormation(llm=llm, max_rules=20), # existing system's component
    utilization=PromptInjection(),
)

# Step 3: Run evaluation with a decoupled protocol
results = BatchProcedural().run(ms, benchmark, EvalSettings(max_test_tasks=10), llm)
\end{lstlisting}
    \caption{A code example of MemTools.}
    \label{fig:code_snippet}
\end{figure}

\section{Demonstration and Framework Evaluation}
We demonstrate the utility of MemTools by addressing the architectural limitations of coupled agent memory designs. The evaluation comprises three validation scenarios: cross-system component integration, the separation of evaluation protocols from datasets, and the coordination of different memory representations.

\subsection{Cross-System Component Integration}
\begin{figure}[t]
\centering
\includegraphics[width=0.85\linewidth]{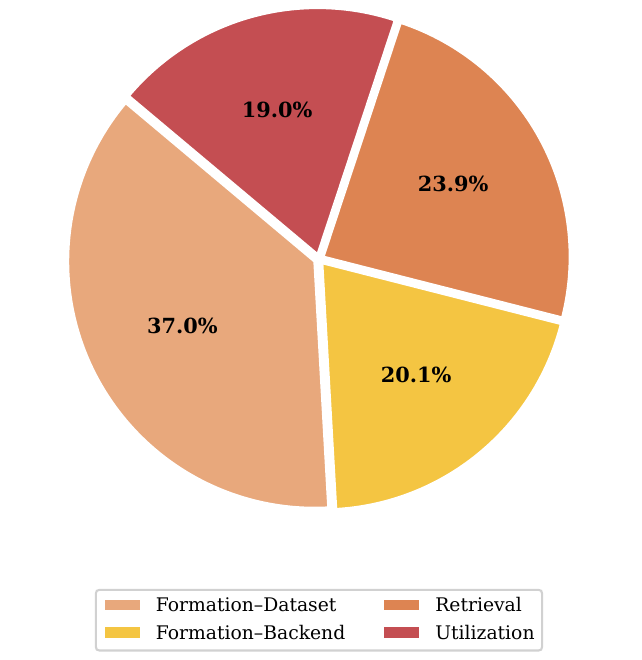}
\caption{Distribution of pipeline incompatibility causes across memory lifecycle stages.}
\label{fig:incompatibility_pie}
\end{figure}

Unaligned data structures present a primary technical barrier when integrating independent memory components. To verify the function of this mechanism, we analyze the distribution of pipeline configuration failures intercepted by the framework during the construction of mixed-architecture systems.
As illustrated in Figure~\ref{fig:incompatibility_pie}, incompatible pipeline assemblies are distributed across the entire memory lifecycle, including dataset-to-formation alignment (37.0\%), memory retrieval (23.9\%), backend storage initialization (20.1\%), and utilization (19.0\%). This distribution indicates that data structural conflicts are not isolated to a specific stage. By actively validating data dependencies across all layers, the matching engine ensures that upstream outputs satisfy downstream requirements before experimental execution begins.

\begin{table}[t]
\centering
\small
\begin{tabular}{llc}
\toprule
\textbf{Pipeline Configuration} & \textbf{Protocol} & \textbf{Success Rate} \\
\midrule
AWM (Native) & Batch & 40.30 \\
AWM (Form.) \\+ A-Mem (Back./Ret.) & Batch &  43.28 \\
AWM (Native) & Stream &  33.58 \\
\bottomrule
\end{tabular}
\caption{Evaluation of memory pipeline configurations on ALFWorld. Rows 1 and 2 compare a native architecture with a cross-system hybrid under the same protocol. Rows 1 and 3 contrast the effects of different evaluation protocols on an identical pipeline.}
\label{tab:pipeline_results}
\end{table}

Beyond structural validation, we evaluate the empirical utility of assembling cross-framework pipelines. We test two distinct memory configurations on ALFWorld~\cite{alfworld} under a unified batch processing protocol.
As reported in Table~\ref{tab:pipeline_results}, a hybrid assembly combining the AWM formation module~\cite{awm} with the backend and retrieval modules from A-Mem~\cite{xu2026amem} achieves a success rate of 43.28, outperforming the native AWM pipeline. This performance variance highlights the practical value of modular interoperability in agent memory research, demonstrating that cross-system combinations can yield performance comparable to or exceeding native setups. MemTools allows researchers to isolate and manipulate this variable systematically.

\subsection{Separating Evaluation Protocols and Datasets}

\begin{figure}[t]
\centering
\includegraphics[width=\linewidth]{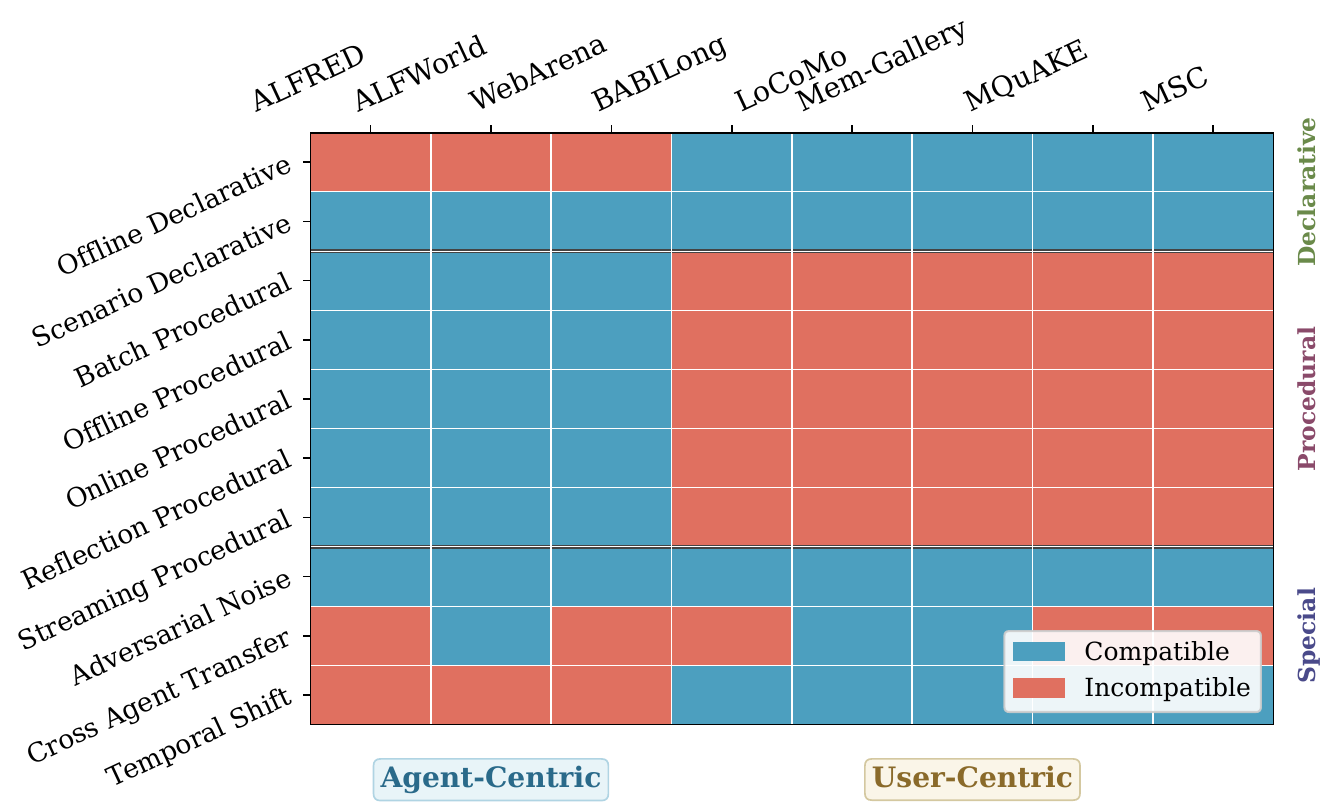} 
\caption{Compatibility heatmap between evaluation protocols and benchmark datasets.}
\label{fig:protocol_heatmap}
\end{figure}

MemTools separates the dataset (acting as a data provider) from the evaluation protocol (dictating execution timing). Figure~\ref{fig:protocol_heatmap} illustrates the compatibility between implemented protocols and benchmarks. The clustering reveals a bifurcation based on data requirements, separating agent-centric environments from user-centric question-answering tasks. This structural divide confirms that protocols operate as independent entities governed by explicit data interfaces rather than arbitrary scripts, justifying their abstraction within the framework.

To demonstrate the empirical necessity of this isolation, we evaluate the identical AWM pipeline on ALFWorld under different execution timings as shown in Table~\ref{tab:pipeline_results}. 
The pipeline achieves a success rate of 40.30 under the Batch protocol, where trajectories are processed collectively. However, under the Stream protocol, which interleaves memory accumulation with ongoing task execution, the success rate drops to 33.58. Since the underlying memory components and benchmark data remain identical, this degradation is entirely attributable to the timing of memory formation. This observation aligns with recent findings indicating that continuous, incremental updates can impair memory utility over time~\cite{zhang2026usefulmemoriesfaultycontinuously}. Without orthogonal protocol decoupling, it would be difficult to isolate whether a performance drop stems from algorithmic flaws in the memory components or from the continuous update mechanism itself. By framing the evaluation protocol as a configurable parameter, MemTools enables researchers to accurately isolate the sources of performance variation.

\begin{figure}[t]
\centering
\includegraphics[width=\linewidth]{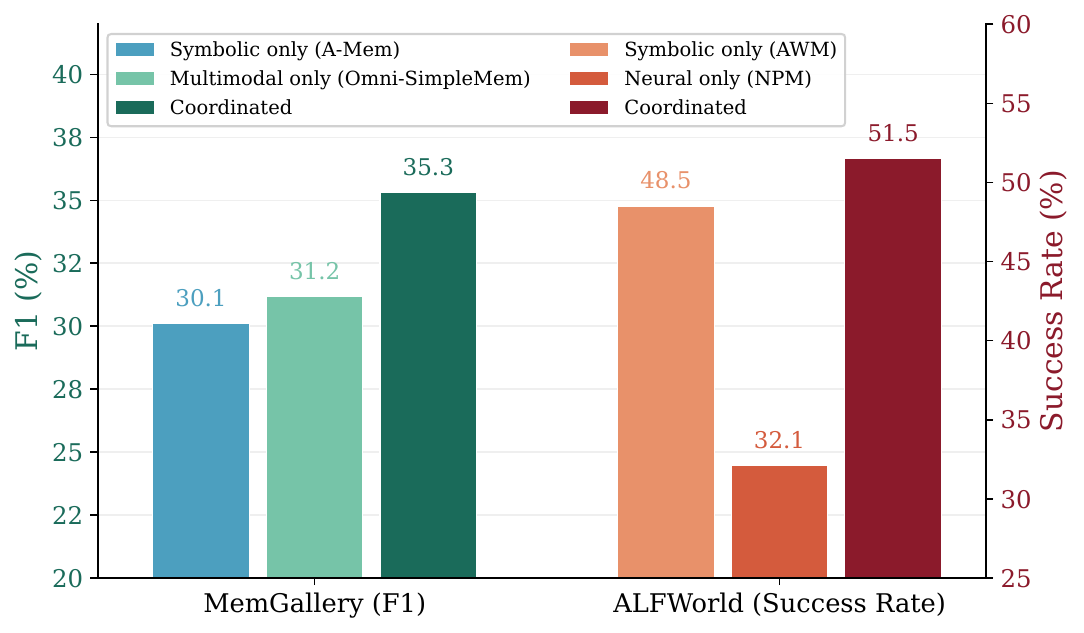}
\caption{Performance comparison of isolated memory representations versus coordinated multi-system pipelines. The left axis reports F1 scores on Mem-Gallery for symbolic and multimodal coordination, while the right axis reports success rates on ALFWorld for symbolic and neural coordination.}
\label{fig:heterogeneous_coordination}
\end{figure}

\subsection{Coordinating Heterogeneous Memory Representations}

Managing heterogeneous memory representations requires coordinating fundamentally different data formats and mechanisms. MemTools handles this by encapsulating each representation into an independent pipeline under a standardized lifecycle interface. 
We evaluate this coordination mechanism by dividing our analysis into two sub-experiments testing pairwise modality integration, addressing the constraint that individual benchmarks rarely demand symbolic, neural, and multimodal memory simultaneously. Figure~\ref{fig:heterogeneous_coordination} presents the performance of isolated memory pipelines against their coordinated configurations. On the Mem-Gallery benchmark~\cite{memgallery}, coordinating the symbolic pipeline A-Mem with the multimodal pipeline Omni-SimpleMem~\cite{omni-simplemem} yields an F1 score of 35.3, surpassing their independent performances of 30.1 and 31.2, respectively. Similarly, on ALFWorld, coordinating the symbolic pipeline AWM with the neural pipeline NPM~\cite{NPM} achieves a 51.5 success rate, outperforming both individual baselines (48.5 and 32.1). These empirical gains confirm that distinct memory representations provide complementary capabilities during task execution. By encapsulating these distinct technologies under a unified interface, MemTools allows researchers to systematically measure the utility of hybrid memory systems.

\section{Conclusion}

This paper introduces MemTools, an interoperability research framework designed to mitigate structural fragmentation in agent memory research. By standardizing the memory lifecycle through declarative data contracts, orthogonally decoupling evaluation protocols from benchmark datasets, and unifying heterogeneous representations under a shared interface, the framework effectively decouples core memory pipelines from their deployment environments. Case studies demonstrate the utility of the framework. Overall, MemTools provides a versatile and robust infrastructure for advancing systematic, comparative research in agent memory.

\section*{Limitations}

While MemTools facilitates interoperability, its automatic matching engine only verifies the structural compatibility of data fields. Subtle system differences may cause behavioral misalignments between components. The abstraction layers required for this modularity also introduce a slight computational overhead during evaluation loops compared to tightly coupled, native systems. Consequently, the current implementation is optimized for controlled research environments and may encounter performance bottlenecks when coordinating massive heterogeneous databases over exceptionally long task horizons.


\bibliography{main}

\appendix
\clearpage
\section{Declarative Data Contracts}

Components in MemTools explicitly declare their input (\textbf{requires}) and output (\textbf{provides}) data fields. During initialization, a matching engine verifies pipeline compatibility by ensuring that upstream outputs satisfy downstream inputs (\verb|provides| $\supseteq$ \verb|requires|). Table~\ref{tab:key_vocabulary} lists the standard keys defining this shared data contract, categorized by their role. This vocabulary is extensible, allowing users to define new keys for custom data modalities or component functions.

A pipeline configuration is only executable when each downstream component's required keys are satisfied by the union of keys provided by its upstream components. For example, a retrieval module requiring \texttt{text\_embedding\_index} only matches with a backend capable of generating dense embeddings. The matching engine automatically enumerates all valid combinations and intercepts incompatible assemblies before the evaluation begins.

\section{Evaluation Protocols and Benchmarks}

Table~\ref{tab:protocols} summarizes the benchmark datasets and evaluation protocols currently supported by MemTools.

\section{Supported Components and Extension Guidelines}

\subsection{Integrated Adapters}

Table~\ref{tab:adapters} details the existing memory systems adapted into the MemTools framework. Each system implements specific components within the memory lifecycle and processes specific memory representations. Users can assemble hybrid pipelines by combining components across different adapters, provided their data contracts match. Additionally, MemTools also provides several built-in components for basic construction.

\subsection{Adding New Components and Adapters}

Researchers can integrate custom components without modifying the core framework by subclassing the relevant base class from \texttt{memtools.core.registry}, such as \texttt{MemoryBackend} or \texttt{RetrievalStrategy}. During this implementation, developers establish the data contract by explicitly assigning the \texttt{requires\_keys} and \texttt{provides\_keys} class attributes. The new module is then placed in the appropriate package directory, like \texttt{memtools/backends/}, where the framework automatically discovers and loads it at runtime using Python's package utilities. Integrating an external memory system follows an identical procedure where the adapter implementation is placed in \texttt{memtools/adapters/}, allowing the matching engine to validate its data keys and recognize it as an available pipeline module.

\subsection{Adding New Protocols and Benchmarks}

Expanding the framework's evaluation capabilities involves a comparable class extension process. Custom evaluation protocols inherit from the \texttt{EvalProtocol} base class. Protocol developers specify necessary dataset dependencies through attributes like \texttt{requires\_sample\_fields} and define the specific execution logic within the \texttt{run} method to output structured \texttt{EvalRecord} objects. Correspondingly, new benchmark datasets inherit from the \texttt{Benchmark} base class. Developers declare the \texttt{provides\_sample\_fields} attribute to expose available data and implement the \texttt{load\_train}, \texttt{load\_test} and \texttt{solve} methods to manage task samples. Before execution begins, the framework ensures structural compatibility by actively matching the benchmark's provided fields against the protocol's required attributes.

\section{User Study}
\label{sec:developer-study}

To assess whether MemTools reduces developer-side integration effort, we recruited three NLP graduate student annotators to complete a small developer study. Each annotator implemented the same composite memory pipeline and recorded two implementation-level metrics: effective lines of code (LOC) and the number of custom functions/classes. LOC counts non-empty, non-comment lines, while Custom Func./Cls. counts developer-defined \texttt{class} and \texttt{def} declarations, excluding imported MemTools components. As shown in Table~\ref{tab:developer-study}, MemTools reduces code size by 60.1\% and custom definitions by 85.4\%.

\begin{table}[t]
\centering
\small
\begin{tabular}{lcc}
\toprule
Condition & LOC $\downarrow$ & Custom Func./Cls. $\downarrow$ \\
\midrule
Native & $135.5 \pm 94.0$ & $24.0 \pm 12.7$ \\
MemTools & $54.0 \pm 4.2$ & $3.5 \pm 3.5$ \\
\midrule
Relative reduction & $60.1\%$ & $85.4\%$ \\
\bottomrule
\end{tabular}
\caption{Developer effort for building a composite memory pipeline.}
\label{tab:developer-study}
\end{table}

\begin{table*}[h]
\centering
\begin{tabular}{@{}lp{10cm}@{}}
\toprule
\textbf{Category} & \textbf{Keys} \\
\midrule
Backend Index &
    \texttt{id},
    \texttt{timestamp},
    \texttt{text\_index},
    \texttt{text\_embedding\_index},
    \texttt{image\_embedding\_index},
    \texttt{audio\_embedding\_index},
    \texttt{video\_embedding\_index},
    $\ldots$ \\
\midrule
MemoryEntry Content &
    \texttt{text\_content},
    \texttt{image\_content},
    \texttt{audio\_content},
    \texttt{video\_content},
    \texttt{kv\_cache\_content},
    \texttt{lora\_adapter\_content},
    \texttt{task\_type},
    $\ldots$ \\
\midrule
Sample Metadata &
    \texttt{sample\_id},
    \texttt{question},
    \texttt{reference},
    \texttt{task\_spec},
    \texttt{success},
    \texttt{steps},
    $\ldots$ \\
\midrule
Raw / Temporal Content &
    \texttt{raw\_text},
    \texttt{raw\_image},
    \texttt{raw\_audio},
    \texttt{raw\_video},
    \texttt{pre\_raw\_text},
    \texttt{pre\_raw\_image},
    \texttt{pre\_raw\_audio},
    \texttt{pre\_raw\_video},
    \texttt{post\_raw\_text},
    \texttt{post\_raw\_image},
    \texttt{post\_raw\_audio},
    \texttt{post\_raw\_video},
    $\ldots$ \\
\bottomrule
\end{tabular}
\caption{Standard key vocabulary categorized by data flow role. The list can be extended for custom requirements.}
\label{tab:key_vocabulary}
\end{table*}

\begin{table*}[h]
\centering
\begin{tabular}{@{}lp{11cm}@{}}
\toprule
\textbf{Category} & \textbf{Items} \\
\midrule
Protocols &
    OfflineDeclarative,
    OfflineProcedural,
    OnlineProcedural,
    BatchProcedural,
    StreamingProcedural,
    AdversarialNoise,
    TemporalShift,
    CrossAgentTransfer,
    ReflectionProcedural,
    ScenarioDeclarative,
    $\ldots$ \\
\midrule
Benchmarks &
    LoCoMo~\cite{locomo},
    MSC~\cite{msc},
    MQuAKE~\cite{zhong-etal-2023-mquake},
    Mem-Gallery~\cite{memgallery},
    ALFWorld~\cite{alfworld},
    ALFRED~\cite{ALFRED20},
    WebArena~\cite{webarena},
    BABILong~\cite{BABILong},
    $\ldots$ \\
\bottomrule
\end{tabular}
\caption{Evaluation protocols and benchmarks supported in MemTools.}
\label{tab:protocols}
\end{table*}

\begin{table*}[h]
\centering
\begin{tabular}{@{}l c c c c c@{}}
\toprule
\textbf{Adapter} & \textbf{Formation} & \textbf{Retrieval} & \textbf{Backend} & \textbf{Evolution} & \textbf{Utilization} \\
\midrule
SimpleMem~\cite{simplemem}      & \ding{108} & \ding{108} & \ding{108} & --- & --- \\
Mem0~\cite{mem0}           & \ding{108} & \ding{108} & \ding{108} & \ding{108} & --- \\
ExpeL~\cite{expel}          & \ding{108} & --- & --- & --- & --- \\
AWM~\cite{awm}            & \ding{108} & \ding{108} & --- & --- & ---  \\
LightMem~\cite{lightmem}       & \ding{108} & \ding{108} & \ding{108} & \ding{108} & ---  \\
StructMem~\cite{structmem}      & \ding{108} & \ding{108} & \ding{108} & \ding{108} & ---  \\
A-Mem~\cite{xu2026amem}          & \ding{108} & \ding{108} & \ding{108} & --- & ---  \\
NPM~\cite{NPM}            & \ding{108} & \ding{108} & \ding{108} & --- & \ding{108}  \\
SELF-PARAM~\cite{wang2025selfupdatable}     & \ding{108} & \ding{108} & \ding{108} & --- & \ding{108} \\
DyPRAG~\cite{dyprag}         & --- & --- & --- & --- & \ding{108} \\
Omni-SimpleMem~\cite{omni-simplemem} & \ding{108} & \ding{108} & \ding{108} & --- & \ding{108} \\
\bottomrule
\end{tabular}
\caption{Adapter components integrated in MemTools. \ding{108} = provided; --- = not provided by this adapter.}
\label{tab:adapters}
\end{table*}

\end{document}